\documentclass[9pt]{extarticle}
\usepackage{spconf,amsmath,graphicx}
\usepackage{comment}
\usepackage{amssymb}
\usepackage{bm}
\usepackage{afterpage}
\usepackage{subfigure}
\title{Leveraging neural Koopman operators to learn continuous representations of dynamical systems from 
scarce
data}

\name{Anthony Frion$^1$, Lucas Drumetz$^1$, Mauro Dalla Mura$^{2,3}$, Guillaume Tochon$^4$, Abdeldjalil A\"{i}ssa-El-Bey$^1$\thanks{This work was supported by Agence Nationale de la Recherche under grant ANR-21-CE48-0005 LEMONADE.}}
\address{$^1$IMT Atlantique, UMR CNRS 6285, Lab-STICC, F-29238 Brest, France\\
$^2$Univ. Grenoble Alpes, CNRS, Grenoble INP, GIPSA-lab, 38000 Grenoble, France\\
$^3$Institut Universitaire de France (IUF), France\\
$^4$LRE EPITA, Le Kremlin-Bicêtre, France}

\begin{document}
%


\maketitle

\begin{abstract}

Over the last few years, several works have proposed deep learning architectures to learn dynamical systems from observation data with no or little knowledge of the underlying physics. A line of work relies on learning representations where the dynamics of the underlying phenomenon can be described by a linear operator, based on the Koopman operator theory. However, despite being able to provide reliable long-term predictions for some dynamical systems in ideal situations, the methods proposed so far have limitations, such as requiring to discretize intrinsically continuous dynamical systems, leading to 
data loss, especially when handling incomplete or sparsely sampled data.
Here, we propose a new deep Koopman framework that represents dynamics in an intrinsically continuous way, leading to better performance on 
limited training data, as exemplified on several datasets arising from dynamical systems.

\end{abstract}
\vspace{-.2cm}
\section{Introduction}
\label{sec:intro}

Learning dynamical systems from observation data with no or only partial prior knowledge of the underlying physical phenomenon has become an important topic in machine and deep learning in recent years~\cite{ayed2019learning}. Many methods have been proposed to 
model dynamical systems
when 
we assume that the data has been generated
from some ordinary differential equation (ODE), usually considering only one training trajectory of the observed system, i.e. the evolution of the system starting from one given initial condition~\cite{foster2020learning,fablet2018bilinear}. 
Doing so
typically makes approaches less robust in inference phase when considering different initial conditions than the one used for training. Moreover, a large number of approaches require a discretization of the system with a fixed time step.
This can be problematic since many dynamical systems, especially related to physical processes, are intrinsically continuous and observations can be irregularly sampled~\cite{brunton2016discovering}. While one can easily design an ideal synthetic dataset with regular high-frequency data (i.e. observations sampled at a high rate), this is not always the case for natural data. For example, optical satellite image time series are generally acquired every several days, and some of the images can be partially or totally unexploitable when the ground is covered by clouds, resulting in spatially and temporally irregularly sampled data~\cite{coluzzi2018first}. Therefore, there is a growing need to be able to handle low-frequency or even irregularly-sampled time series data. Models handling intrinsically continuous dynamics can naturally answer this issue, contrary to recent approaches that need to 
re-sample irregular data on a fixed discrete grid~\cite{brajard2020combining,nguyen2019like}.

\vspace{-.2cm}
\section{Relation to prior work}
\label{SOTA}

Our approach is inspired by the Koopman operator theory~\cite{koopman1931hamiltonian, koopman1932dynamical}. We assume that the state of our dynamics is described by a vector $\mathbf{x} \in X \subseteq \mathbb{R}^n$, $n$ being the dimension of the system. The dynamics can then be modeled by a discrete operator $F: X \to X$ (assumed autonomous here) such that 
\begin{equation}
\label{dynamics}
F(\mathbf{x}_t) = \mathbf{x}_{t+1}
\end{equation}
where we write $\mathbf{x}_t = \mathbf{x}(t)$ at an integer time $t$ for simplicity.
The linear Koopman operator $\mathcal{K}$, which can be applied to any given measurement function $g : X \to \mathbb{R}$ of the dynamical system, consists in composing functions by a time increment:
\begin{equation}
    \mathcal{K}g(\mathbf{x}_t) = g(F(\mathbf{x}_t)) = g(\mathbf{x}_{t+1}).
\end{equation}
The Koopman operator $\mathcal{K}$ is linear, which makes it very powerful since it enables to
use linear algebra manipulations even for nonlinear systems. Notably, one can explicitly find eigenfunctions of $\mathcal{K}$,
i.e. for a measurement function $f$ which is a Koopman eigenfuction with eigenvalue $\lambda$, one has, for any time $t$, $\mathcal{K}f(\mathbf{x}_t) = \lambda f(\mathbf{x}_t) = f(\mathbf{x}_{t+1})$. This offers much more interpretability than a purely nonlinear modeling.
However, for most nonlinear systems, one needs the Koopman eigendecomposition to be infinite-dimensional to exactly model the dynamics. A line of work, starting from \cite{mezic2005spectral} which introduces Koopman mode decomposition, has focused on finding finite-dimensional approximations of the Koopman operator. 
Notably,~\cite{rowley2009spectral} has evidenced a link between the Koopman mode decomposition and the dynamic mode decomposition algorithm~\cite{schmid2009dynamic,schmid2010dynamic}.
The reader is referred to~\cite{brunton2021modern} for a detailed survey on those modern developments of the Koopman operator theory.

Recently, several works have addressed the usage of deep learning architectures in relation with the Koopman operator theory~\cite{takeishi2017learning,morton2018deep,lusch2018deep,li2019learning,otto2019linearly,azencot2020forecasting}. Some of those \cite{azencot2020forecasting,mamakoukas2020learning} have discussed practical ways to promote the long-term stability of the approximated operator. Most of these works do not aim at learning the full Koopman operator, but rather its restriction to a finite set of functions, which is a Koopman Invariant Subspace (KIS)~\cite{takeishi2017learning}. Since such a set is stable by application of the Koopman operator, its restriction to the set can be written as a matrix $\mathbf{K} \in \mathbb{R}^{d\times d}$, $d$ being the dimension of the KIS. Thus,~\cite{takeishi2017learning,otto2019linearly,li2019learning,azencot2020forecasting} try to jointly learn a set of functions that form a KIS and the restriction $\mathbf{K}$ of the Koopman operator to this KIS. The KIS is typically learned by a neural autoencoding network, which means that it has to be informative enough to reconstruct the observation state. The present work also follows this line. 

\cite{lusch2018deep} chooses an alternative approach: they also train an autoencoder on the observation state yet they do not assume that it learns a KIS and they rather try to model a continuous Koopman spectrum. Therefore, they replace the matrix $\mathbf{K}$ by an auxiliary network which takes a latent state as input and returns a matrix to multiply to it to get the next state.


An important aspect of the Koopman framework is that it offers tools to constrain the Koopman latent representation through spectral properties of the linear operator, making learned models more robust e.g. to changes in initial conditions. More importantly, in this paper, we exploit the fact that working with a finite-dimensional linear latent representation allows to propagate the dynamics in the Koopman space to any time instant using a closed form integration of the dynamics.
Thus, the latent dynamics are essentially continuous since we can query the state of the dynamical system for any instant, starting from a known initial condition. This 
enables a very natural and effective way to 
perform predictions that are not bound to be discrete with the same sampling frequency as the training data.


\vspace{-.2cm}
\section{Proposed method}
\label{sec:method}

\subsection{The discrete setting}

In this setting, we use the same assumptions and notations as in equation~\eqref{dynamics}.
Our dataset is composed of $m$ discrete trajectories of T regularly sampled points.
Given the $m$ initial conditions $\mathbf{x}_{i,0}$ corresponding to each trajectory with index $1 \leq i \leq m$, one can write:
\begin{equation}
  \quad \forall \ 1 \leq i \leq m, \forall \ 0 \leq t \leq T, \quad \mathbf{x}_{i, t} = F^t(\mathbf{x}_{i,0}).
\end{equation}
The key to being able to represent the data with a finite-dimensional linear operator is to find a Koopman invariant subspace \cite{takeishi2017learning} (KIS),
i.e. a set $J$ of measurement functions such that $\forall g \in J, \quad \mathcal{K}(g) \in J$. While there exists trivial KIS, for example with observable functions corresponding to constant values, we need to find one that contains enough information to reconstruct the observation state $\mathbf{x}$.

Our chosen architecture is composed of 3 trainable components: 
(1) a deep encoder network $\phi : \mathbb{R}^n \to \mathbb{R}^d$, (2) the corresponding decoder network $\psi : \mathbb{R}^d \to \mathbb{R}^n$ and (3) a square matrix $\mathbf{K} \in \mathbb{R}^{d\times d}$.
These components are the same as in the Linearly-Recurrent Autoencoder Network (LRAN) from \cite{otto2019linearly}. The idea is that $\phi$ and $\psi$ represent a nonlinear mapping from the space of observables to a KIS and vice versa, while $\mathbf{K}$ corresponds to the restriction of the Koopman operator of the system to this KIS.
In theory, these components should allow us to make predictions with:
\begin{equation}
    \forall \ 1\leq i\leq m, \forall \ 0\leq t_0\leq t_1 \leq T, \mathbf{x}_{i,t_1} \simeq \psi(\mathbf{K}^{t_1 - t_0}\phi(\mathbf{x}_{i,t_0})). 
\end{equation}

In order to favor such behavior, we train our architecture using compositions of the
following generic loss function terms:
\begin{align}
\label{mid_term}
\begin{aligned}L_{pred, \Delta t}(\phi, \psi, \mathbf{K}, i, t) =  ||\mathbf{x}_{i,t+\Delta t} - \psi(\mathbf{K}^{\Delta t}\phi(\mathbf{x}_{i,t}))||^2 \\
L_{lin, \Delta t}(\phi, \psi, \mathbf{K}, i, t) = ||\phi(\mathbf{x}_{i,t+\Delta t}) - \mathbf{K}^{\Delta t}\phi(\mathbf{x}_{i,t})||^2 \end{aligned}
\end{align}
where $t$ and $\Delta t$ are integers.
$L_{pred}$ and $L_{lin}$ correspond respectively to prediction and linearity loss terms. 
While the prediction loss directly favors the accuracy of predictions, the linearity loss rather ensures a coherent latent trajectory.
Note that
$L_{pred,0}(\phi, \psi, \mathbf{K}, i, t) = ||\mathbf{x}_{i,t} - \psi(\phi(\mathbf{x}_{i,t}))||^2$ is an autoencoding loss term, which is called the reconstruction loss in~\cite{lusch2018deep}, and that $L_{lin,0}(\phi, \psi, \mathbf{K}, i, t) = 0$.

The authors from \cite{otto2019linearly} use a long-term loss function which we rewrite here as 
\begin{multline}
\label{long_term}
L(\phi, \psi, \mathbf{K}) = \sum_{1 \leq i\leq m, \ 0 \leq t \leq T-1} \delta^t L_{pred,t}(\phi, \psi, \mathbf{K}, i, 0) \\ 
+ \beta \delta^t L_{lin,t}(\phi, \psi, \mathbf{K}, i, 0) + \Omega(\phi, \psi, \mathbf{K})
\end{multline}
where $\beta$ determines the relative importance of the prediction and linearity terms, $\Omega$ is a regularization term (included for generality though the method from~\cite{otto2019linearly} does not use any) and $\delta < 1$ is used to give more importance to the earliest time steps, making the optimization easier.
In our own long-term loss function~\eqref{L2}, we use $\beta = 1$, keep the same weight for all time steps (i.e. set $\delta=1$), and add the reconstruction loss for all samples. 

A novelty of our work is that we softly enforce that $\mathbf{K}$ belongs to the orthogonal group, which means that it defines a map close to a Euclidean isometry in $\mathbb{R}^{d}$, and as such approximately preserves norms in the latent space. Indeed, assuming that $\mathbf{K}$ is not orthogonal would necessarily involve one of the two following options:

(1) Some of the eigenvalues have a modulus strictly greater than one. Then the operator norm $|||\mathbf{K}|||$ of $\mathbf{K}$ (the maximum singular value of $\mathbf{K}$) would be such that $|||\mathbf{K}|||> 1$. Since 
$\forall t<T, ||\mathbf{K}^t\phi(\mathbf{x}_{0})|| \leq |||\mathbf{K}|||^{t} ||\phi(\mathbf{x}_{0})||$, 
for some initial conditions (e.g. when $\phi(\mathbf{x}_{0})$ is an eigenvector corresponding to the largest eigenvalue of $\mathbf{K}$, i.e. when the bound is tight), the system may quickly reach a region of the latent space that has never been observed during training.


(2) Some of the eigenvalues have a modulus strictly smaller than one. Then, for any initial condition, the latent state will converge to a subspace of lower dimension spanned by eigenvectors corresponding to the remaining eigenvalues. While this behavior makes sense for systems that converge to a low-dimensional limit cycle, it can be problematic when modeling 
a conservative dynamical system

Thus, the orthogonality of $\mathbf{K}$ enables to make stable predictions on a far longer horizon. The Koopman operator framework could allow us to enforce this orthogonality constraint exactly, by forcing the eigenvalues of $\mathbf{K}$ to lie on the unit circle. However, we choose to favor orthogonality in a soft way, through a loss term 
which was previously discussed in the litterature (e.g. by \cite{vorontsov2017orthogonality} and \cite{bansal2018can}), and 
which can be expressed as 
\begin{equation}
\label{orthogonality}
L_{orth}(\mathbf{K}) = ||\mathbf{KK}^T - \mathbf{I}||_{F}^2
\end{equation}
where $\mathbf{I}$ is the identity matrix and $||\cdot||_F$ the Frobenius norm. This soft penalization is in practice more flexible than a hard constraint. 

To sum up, our discrete models are first trained with a short-term loss $L_1$ and then re-trained with a long-term loss $L_2$, which are expressed as follows:
\begin{multline}
\label{L1}
    L_1(\phi, \psi, \mathbf{K}) = \beta_1 L_{orth}(\mathbf{K}) + \sum_{1 \leq i\leq m, \ 0 \leq t \leq T-5} L_{pred,0}(i, t) \\
    + L_{pred,1}(i, t) + L_{pred,5}(i, t) + L_{lin,1}(i, t) + L_{lin,5}(i,t)
\end{multline}
\begin{multline}
\label{L2}
    L_2(\phi, \psi, \mathbf{K}) = \beta_2 L_{orth}(\mathbf{K}) 
    + \sum_{1 \leq i\leq m, \ 0 \leq t \leq T-1} L_{pred,0}(i, t) \\
    + L_{pred,t}(i, 0)
    + L_{lin,t}(i, 0)
\end{multline}

where we dropped the dependencies of $L_{pred}$ and $L_{lin}$ on $\mathbf{K}$, $\phi$ and $\psi$ for clarity.
Note that we did not observe a need for setting relative weights between the different prediction and linearity loss terms, while the regularizating orthogonality terms most often require high weights $\beta_1$ and $\beta_2$ to have a practical influence on the training. Also, $L_1$ is an arbitrary combination of terms which proved effective in most of our experiments, yet one could easily construct another similar loss using different terms from~\eqref{mid_term}, the most important one probably being the reconstruction loss $L_{pred,0}$.
\vspace{-.2cm}
\subsection{Generalizing a discrete model to a continuous setting}
\label{discrete_to_continuous}

Let us assume that we are modeling a continuous dynamical system and that, inside of the latent space spanned by the previously described encoder $\phi$, there exists a linear infinitesimal operator, represented by a matrix $\mathbf{D}$, such that:

\begin{equation}
\label{infinitesimal}
    \frac{d\phi(\mathbf{x})}{dt} = \frac{d\mathbf{z}}{dt} = \mathbf{D}\mathbf{z}.
\end{equation}
By integrating this infinitesimal operator, one is now able to advance time by any desired amount, not necessarily on a discrete and regular grid. In particular, knowing the initial condition of the system $\mathbf{z}(t_0) = \phi(\mathbf{x}(t_0))$, one can write:
\begin{equation}
\label{integration}
    \mathbf{z}(t_0+\Delta t) = \exp(\Delta t\mathbf{D})\mathbf{z}(t_0)
\end{equation}
where $\exp$ stands for the matrix exponential.

In particular, 
with $\Delta t = 1$, one can identify that 
\begin{equation}
\label{K-D}
\mathbf{K} = \exp(\mathbf{D}).
\end{equation}
Such a real matrix $\mathbf{D}$ exists under technical conditions on the Jordan normal form of $\mathbf{K}$~\cite{culver1966existence}, which may only fail when $\mathbf{K}$ has real negative eigenvalues (close to $-1$, since this is the only possible negative eigenvalue for an orthogonal matrix). This is unlikely to happen in practice and never occurred in our experiments\footnote{Should we constrain $\mathbf{K}$ to be perfectly orthogonal, a real logarithm would always exist since $O(d)$ is a matrix Lie group, whose Lie algebra is the set of skew-symmetric matrices~\cite{baker2012matrix}}.


To obtain a matrix logarithm efficiently, let us write the eigendecomposition of $\mathbf{K}$ in $\mathbb{C}$:
\begin{equation}
    \mathbf{K} = \mathbf{V} \mathbf{\Lambda} \mathbf{V}^{-1}
\end{equation}
where $\mathbf{V}\in \mathbb{C}^{d \times d }$ and $\mathbf{\Lambda} \in \mathbb{C}^{d \times d }$ is a diagonal matrix containing the (complex) eigenvalues of $\mathbf{K}$. Then we can efficiently compute $\mathbf{D}$ by taking the principal logarithm of each (necessarily not real negative) eigenvalue:

\vspace{-0.2cm}
\begin{equation}
    \mathbf{D} = \mathbf{V}\log(\mathbf{\Lambda})\mathbf{V}^{-1}
\end{equation}
where $\log(\mathbf{\Lambda})$ means the principal logarithm is applied to each diagonal element of $\mathbf{\Lambda}$, and the off diagonal elements are $0$. 

\vspace{-.2cm}
\section{Experiments}
\label{sec:exp}
We perform experiments on three dynamical systems: the simple pendulum, a 3-dimensional dynamical system modeling the low-dimensional attractor of a nonlinear high-dimensional fluid flow, and the chaotic Lorenz-63 system. The first two benchmarks are taken from \cite{lusch2018deep} and we refer the reader to this paper for further detail. The Lorenz system is a well-known 3-dimensional chaotic system with a strange attractor, driven by the equations:
\begin{equation}
    \frac{dx}{dt} = \sigma(y-x) ; \quad \frac{dy}{dt} = x(r-z)-y ; \quad \frac{dz}{dt} = xy-bz
\end{equation}
with parameters $\sigma = 10, r=28, b=8/3$.

\begin{figure}
    \centering
    \includegraphics[width=8.5cm]{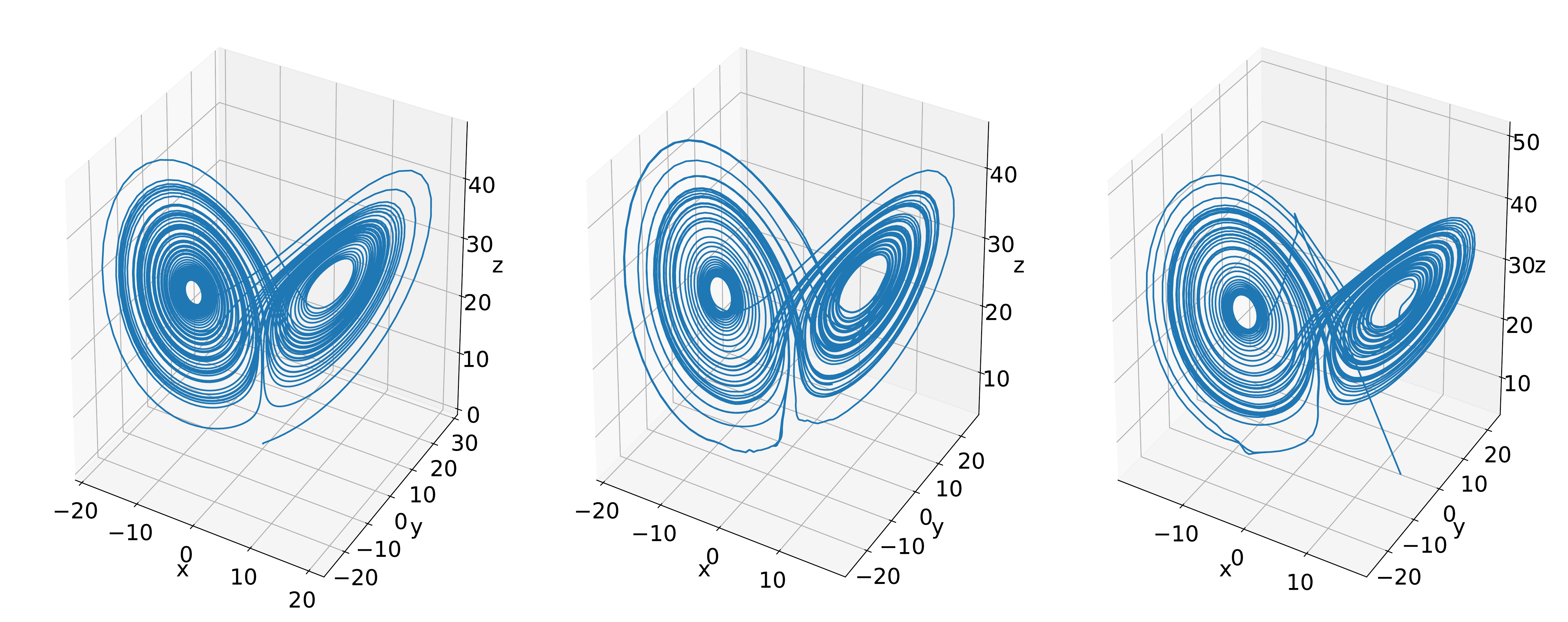}
    \caption{Left: true Lorenz-63 system. Middle: our prediction starting from a point in the attractor. Right: our prediction starting from a point outside of the attractor.}
    \label{fig:Lorenz-63}
\end{figure}

All of our models evaluated in this section have the same architecture: the encoder network $\phi$ is a multi-layer perceptron (MLP) with 3 layers, from which the hidden sizes are 256 and 128, with ReLU nonlinearities and a final Koopman embedding of dimension $d=16$. The decoder also is a 3-layer MLP with symmetric hidden sizes of 256 and 128. Only the size of the observation space varies: it is 6 for the Lorenz-63 systems (the 3 variables and their derivatives), 2 for the pendulum (the angle $\theta$ and its derivative) and 3 for the fluid flow (the basic 3 variables of the system). The cumulated number of parameters for $\phi$, $\psi$ and $\mathbf{K}$ is about 70 000.

As for the Lorenz-63 system, 
despite the simplicity of our architecture, it is still able to learn to behave in a chaotic way, or at least to simulate a chaotic behavior, even when the initialization is outside of the attractor. We obtain Lyapunov exponents $0.61, 0.07, -14.34$ for simulated trajectories of our model, which is close to the estimated real exponents $0.9056, 0, -14.5721$ of the Lorenz system \cite{sparrow2012lorenz}. We also manage to model the attractor when initializing our predictions outside of it. Note that, while the method from \cite{lusch2018deep} is also able to model the Lorenz-63 dynamics faithfully, classical models like recurrent neural networks are known to fail at this task. We plot 
our simulated trajectories
in figure \ref{fig:Lorenz-63}.

For the pendulum dynamics, we use trajectories of 10 seconds for training and validation. At the highest sampling frequency, 
this corresponds to time series of length 1000. The trajectories all start from a random initial angle with zero initial speed. There are 100 training trajectories and 25 validation trajectories, from which the starting angles are sampled uniformly 
in a wide range of angles. The models are then tested on a test set containing 10 trajectories
also covering a wide range of initial angles.

For the fluid flow dynamics, we use the exact same data as in \cite{lusch2018deep}, with time series of length 121. 
This time, we did not keep any of the training data for validation since we did not observe any overfitting in our experiments.

\begin{figure*}
    
        \centering
        \begin{subfigure}
        
            \includegraphics[width=8.6cm]{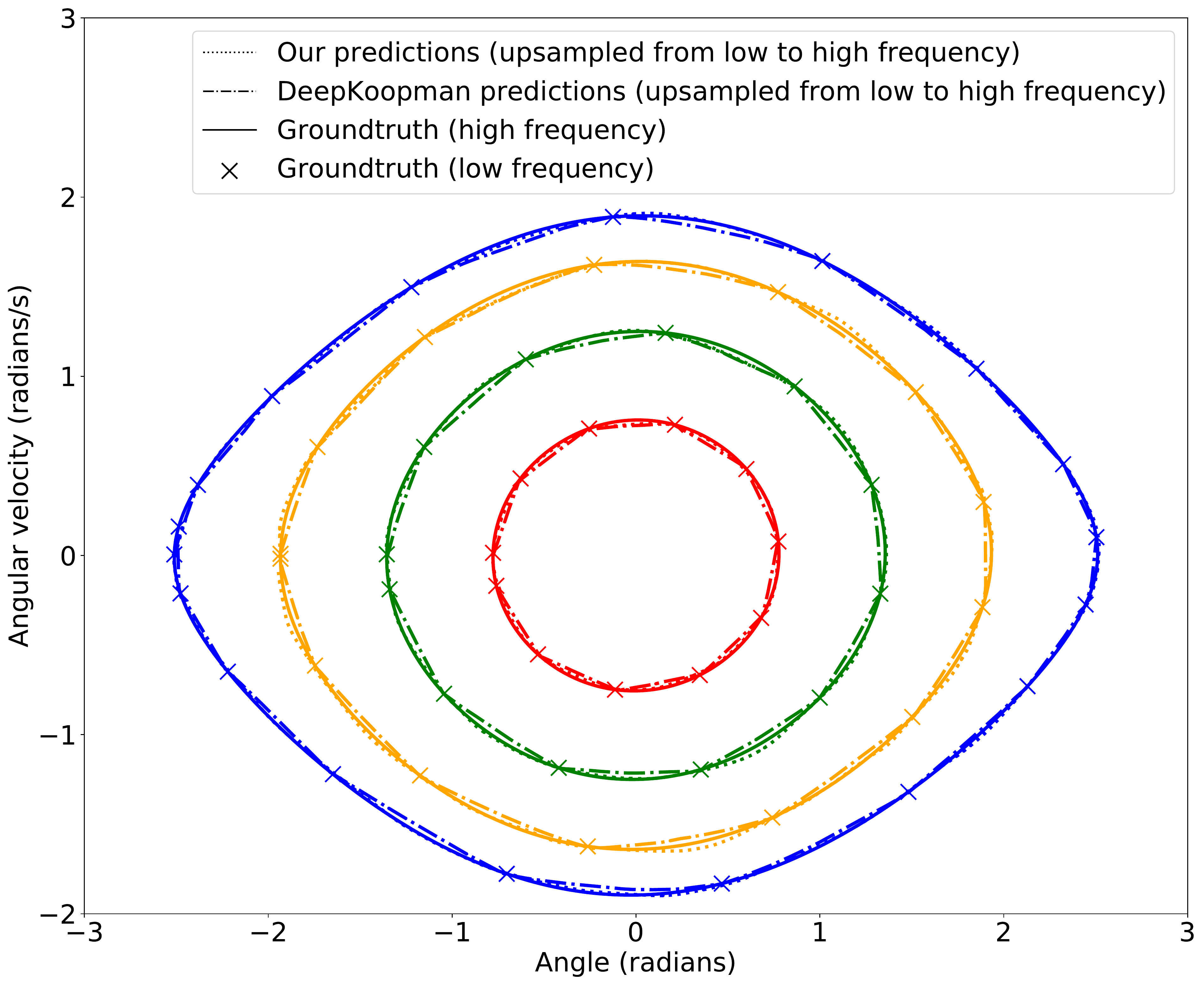}
        \end{subfigure}
        \begin{subfigure}
        
            \includegraphics[width=8.6cm]{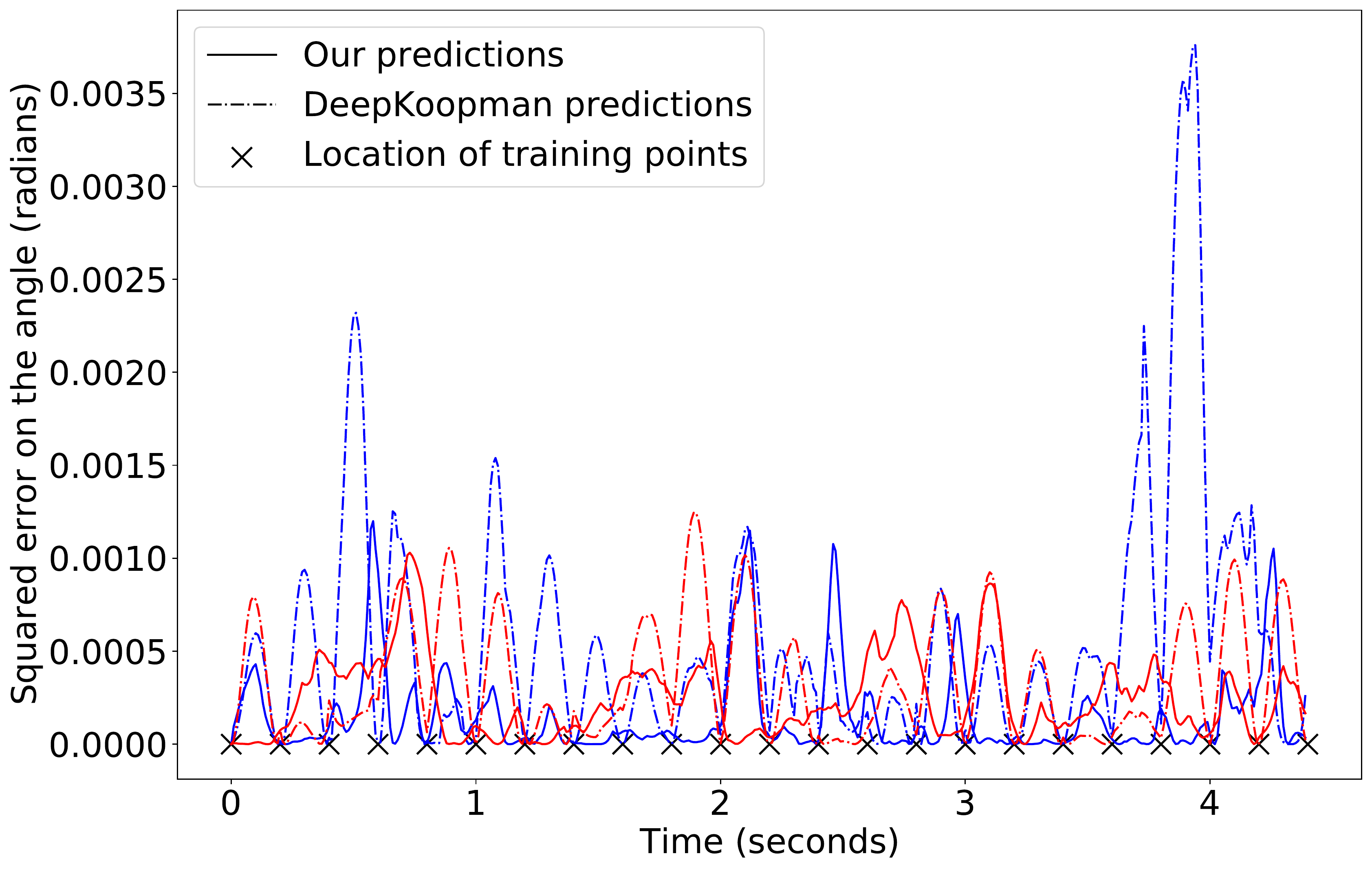}
        \end{subfigure}
        \begin{subfigure}
            \centering
            \includegraphics[width=17.5cm]{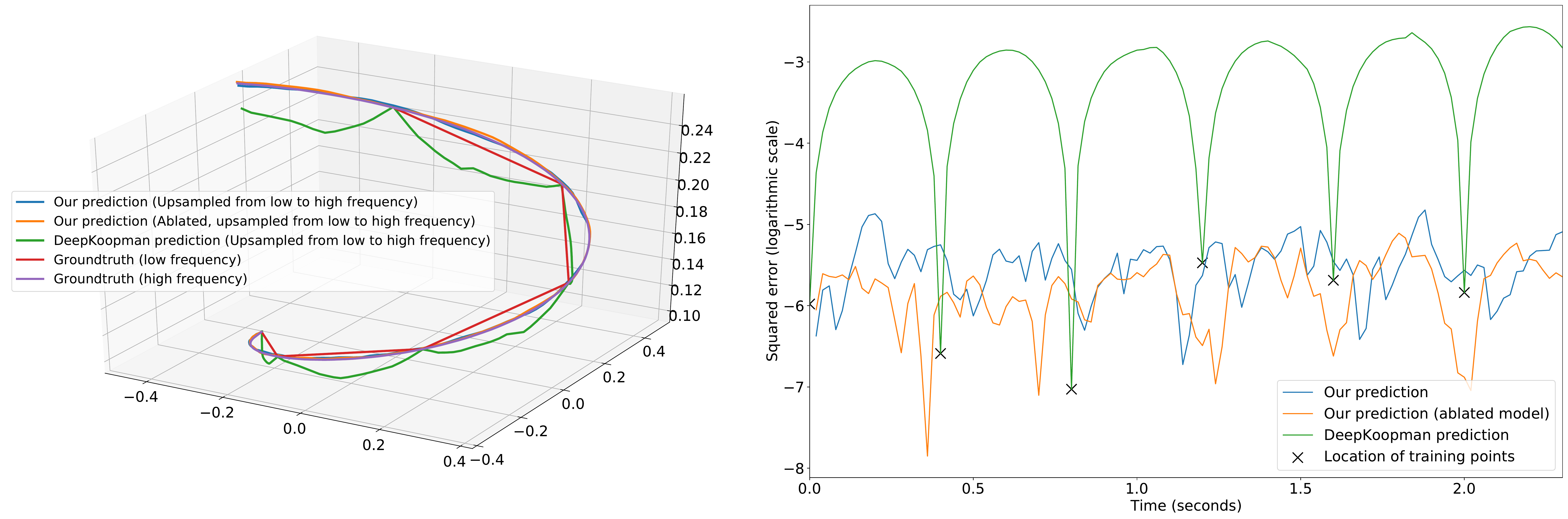}
        
        \end{subfigure}
\caption[size=9pt]{Interpolation results. Top: pendulum, each color represents a different trajectory ; Bottom: fluid flow. Top left: comparison of our predictions against DeepKoopman 
for 4 different testing trajectories. Top right: squared errors between these predictions and the groundtruth. Bottom left: 
upsampled predictions from our model (blue), our ablated model (orange) and DeepKoopman (green)
along with the low-frequency (red) and high-frequency (magenta) sampling of a groundtruth fluid flow trajectory. Bottom right: The corresponding squared errors from the 3 models, in $log_{10}$ scale.}
\label{fig:interpolation}
\end{figure*}

Our evaluation criterion for the pendulum and fluid datasets is the mean squared error, averaged over 
all time steps from all of the testing trajectories. 
We chose this metric since, as one can see on figure \ref{fig:interpolation}, the prediction error 
follows a complex pattern which does not always increase with time.
Thus, the error measured only at one given time step is not representative of the 
accuracy of a prediction. 
We compare our method against the state-of-the art \cite{lusch2018deep}, which we refer to as DeepKoopman, and against an ablated version of our work where we remove the orthogonality loss~\eqref{orthogonality}, i.e. set $\beta_1 = \beta_2 = 0$ in loss functions $L_1$~\eqref{L1} and $L_2$~\eqref{L2}, making the method similar to \cite{otto2019linearly}.

We evaluate two settings: 
the first one is a typical high-frequency regular sampling setting, which was used in \cite{lusch2018deep}. The second one has an increased data sampling period: from 0.01s to 0.2s for the pendulum, and from 0.02s to 0.4s for the fluid flow. This reduced frequency results in an important gap between consecutive sampled points, giving an incomplete view of the dynamical systems.
We evaluate the trained models in the original high frequency in order to assess their ability to learn a continuous evolution from low-frequency data. For our method, we compute a continuous operator $\mathbf{D}$ from the learned matrix $\mathbf{K}$ to eventually provide high-frequency predictions, as described in section~\ref{discrete_to_continuous}. For the method from \cite{lusch2018deep}, since it cannot be used to make predictions at a different frequency from the one it has been trained on, we will simply perform a linear interpolation between the points predicted at the low training frequency. We interpolate in the latent space since it gives slightly better results than interpolating directly in the observation space.




\begin{table}
\centering
{\footnotesize


\begin{tabular}{|c|c|c|c|}
\hline
Dataset & Our method & DeepKoopman & Our method (ablated) \\
\hline
Pendulum (HF) & $5.38\times10^{-4}$ & $\bm{1.41\times10^{-4}}$ & $4.42\times10^{-2}$ \\
\hline
Pendulum (LF) & $\bm{6.82\times10^{-4}}$ & $1.62\times10^{-3}$ & $8.02\times10^{-4}$ \\
\hline
Fluid flow (HF) & $5.50\times10^{-6}$ & $1.21\times10^{-6}$ & $\bm{7.75\times10^{-7}}$ \\
\hline
Fluid flow (LF) & $8.89\times10^{-6}$ & $2.31\times10^{-3}$ & $\bm{2.31\times10^{-6}}$ \\
\hline

\end{tabular}
}

\caption{Mean squared error averaged over all 
points
from all high-frequency testing trajectories. (HF) and (LF) stand respectively for "high frequency" and "low frequency" settings.}
\end{table}
As expected, the original DeepKoopman framework is more accurate than ours in some ideal settings. This is due to the auxiliary network which enables to apply different linear transformations to different latent states while our framework always applies the same transformation. However, our framework outperforms this method in low frequency settings, mostly because it is intrinsically able to upsample its prediction to a higher frequency, resulting in a nontrivial interpolation which is much better than the agnostic linear interpolation. One can visually assess the quality of the interpolation
for the compared models
on figure~\ref{fig:interpolation}.


The ablated version of our work notoriously fails at modeling the pendulum system in high frequency, which shows that the orthogonality constraint is crucial to keep the predictions stable when modeling very long time series, in particular for conservative models such as this one. However, on the fluid flow dataset, the ablated version performs better since the time series are shorter and the system converges to the same limit cycle no matter the initial conditions, so that the matrix $\mathbf{K}$ does not need to be orthogonal for this dataset.

\vspace{-.2cm}
\section{Conclusion}
\label{sec:ccl}

In this paper, we proposed a deep learning implementation of the Koopman operator based on autoencoders so as to construct a linear infinitesimal operator, which enables a natural continuous formulation of dynamical systems.
The infinitesimal operator is constructed from a discrete evolution matrix from which we softly enforce the orthogonality. This enables our model to produce stable predictions
on a far longer time span than previous similar work~\cite{otto2019linearly}, on a time horizon comparable to methods that use an auxiliary network to build a new matrix at each time step~\cite{lusch2018deep}. 
Using our continuous formulation, one can easily interpolate low-frequency data in time to obtain a convincing high-frequency prediction which
is close to the real dynamical system.

Possible extensions of our work include adapting it to controlled dynamical systems, as already proposed by works close to ours~\cite{li2019learning}, and training our architecture jointly to forward and backward prediction for more robustness~\cite{morton2018deep, azencot2020forecasting}.
We have already conducted promising early experiments on learning from irregular data, which are not reported here. This paves the way for future application on difficult natural data, like satellite image time series which are naturally sparse and irregular due to the presence of clouds~\cite{coluzzi2018first}.




\end{document}